\documentclass[11pt,a4paper]{article}
\usepackage[hyperref]{acl2021}
\usepackage{times}
\usepackage{latexsym}

\usepackage{xcolor}
\usepackage{booktabs}
\usepackage{multirow}

\usepackage{graphics}
\usepackage[demo]{graphicx}% so it makes black blobs
\graphicspath{ {plots/} }
\usepackage{microtype}
\usepackage{xurl}
\aclfinalcopy % Uncomment this line for the final submission
\begin{document}

\newcommand\BibTeX{B\textsc{ib}\TeX}
\newcommand{\RomanNumeralCaps}[1]{\MakeUppercase{\romannumeral #1}}
% \DeclareRobustCommand{\hlcyan}[1]{{\sethlcolor{cyan}\hl{#1}}}

\newcommand{\elron}[1]{\textcolor{red}{ELRON: #1}}

% \DeclareRobustCommand{\hlpink}[1]{{\sethlcolor{pink}\hl{#1}}}

\newcommand{\liat}[1]{\textcolor{blue}{LIAT: #1}}
\newcommand{\ranit}[1]{\textcolor{purple}{RANIT: #1}}

\newcommand{\michal}[1]{\textcolor{brown}{MICHAL: #1}}

\newcommand{\QCPGPoint}[0]{\textsc{$QCPG^\star$}}

\title{Quality Controlled Paraphrase Generation}
\author{Elron Bandel$^{1,2}$ \qquad Ranit Aharonov$^1$ \qquad Michal Shmueli-Scheuer$^1$  \\ \textbf{Ilya Shnayderman$^1$ \qquad Noam Slonim$^1$ \qquad Liat Ein-Dor$^1$} \\
  $^1$IBM Research \\
  $^2$Computer Science Department, Bar Ilan University \\
  \texttt{elron.bandel@ibm.com} \\
  \texttt{\{shmueli,ilyashn,noams,liate\}@il.ibm.com}}
\maketitle
\begin{abstract}
Paraphrase generation has been widely used in various downstream tasks.
Most tasks benefit mainly from high quality paraphrases, namely those that are semantically similar to, yet linguistically diverse from, the original sentence.
Generating high-quality paraphrases is challenging as it becomes increasingly hard to preserve meaning as linguistic diversity increases.
Recent works achieve nice results by controlling specific aspects of the paraphrase, such as its syntactic tree. However, they do not allow to directly control the quality of the generated paraphrase, and suffer from low flexibility and scalability.
Here we propose \textsc{QCPG}, a quality-guided controlled paraphrase generation model, that allows directly controlling the quality dimensions.
Furthermore, we suggest a method that given a sentence, identifies points in the quality control space that are expected to yield optimal generated paraphrases. We show that our method is able to generate paraphrases which maintain the original meaning while achieving higher diversity than the uncontrolled baseline. 
The models, the code, and the data can be found in \url{https://github.com/IBM/quality-controlled-paraphrase-generation}.
\end{abstract}
\section{Introduction}

Paraphrase generation, namely rewriting a sentence using different words and/or syntax while preserving its meaning \cite{bhagat2013paraphrase}, is an important technique in natural language processing, that has been widely used in various downstream tasks including question answering \cite{Fader2014QA,mcCann-QA}, summarization \cite{rush-etal-2015-neural-summ}, data augmentation \cite{yu2018qanet} and adversarial learning \cite{iyyer-etal-2018-adversarial-aug}. 
However, not all paraphrases are equally useful. 
For most real-world applications, paraphrases which are too similar to
the original sentence are of limited value, while 
those with high linguistic diversity, i.e. with large syntactic/lexical differences between the paraphrase and the original sentence,
are more beneficial to the robustness and accuracy of automatic text evaluation and classification, and can avoid the blandness caused by repetitive patterns \cite{qian2019exploring}.
The quality of paraphrases is often evaluated using three dimensions, where high quality paraphrases are those with high semantic similarity as well as high lexical and/or syntactic diversity \cite{McCarthy2009TheCO}.

\begin{figure}
\includegraphics[width=\columnwidth]{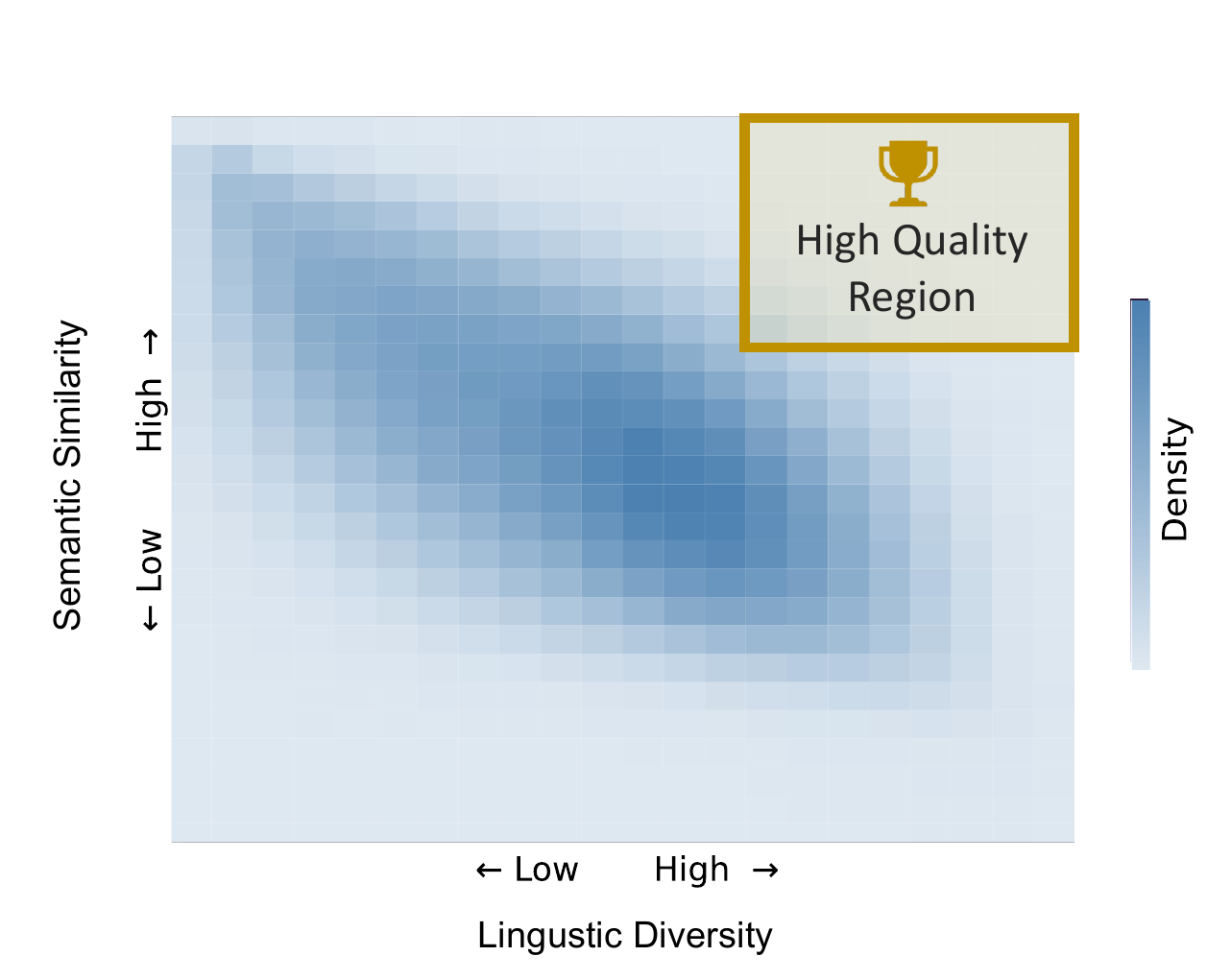}
\caption{Density of paraphrases in WikiAnswers as a function of the semantic similarity and the linguistic diversity. The marked area, which contains high quality paraphrases, is very sparse (The measures used in the figure are described in Section~\ref{quality metrics}) .}
\label{fig:data_density_plot}
\end{figure}

Generating high quality paraphrases can be challenging (for both humans and automatic models) since it is increasingly difficult to preserve meaning with increasing linguistic diversity. 
Indeed, when examining the quality of paraphrases among paraphrase generation datasets, 
one can find a wide range of paraphrase qualities, where the area of high quality is often very sparse (see Figure \ref{fig:data_density_plot}). 
This in turn results in scarcity of supervised data for high-quality paraphrase generation. 

A recent approach aiming to produce high quality paraphrases is controlled paraphrase generation, which exposes control mechanisms that can be manipulated to produce diversity. 
While the controlled generation approaches have yielded impressive results, they require providing the model with very specific information regarding the target sentence, such as its parse tree \cite{iyyer-etal-2018-adversarial-aug} or the list of keywords it needs to contain \cite{zeng2019user}. 
However, for most downstream applications, the important property of the paraphrase is its overall quality, rather than its specific syntactic or lexical form. 
The over-specificity of existing control-based methods not only complicates their usage and limits their scalability, but also hinders their coverage. Thus, it would be desirable to develop a paraphrase generation model, which uses a simple mechanism for directly controlling paraphrase quality, while avoiding unnecessary complications associated with fine-grained controls. 

In this paper we propose \textsc{QCPG}, a Quality Controlled Paraphrase Generation model, that given an input sentence and quality constraints, represented by a three dimensional vector of semantic similarity, and syntactic and lexical distances, produces a target sentence that conforms to the quality constraints. 

Our constraints are much simpler than previously suggested ones, such as parse trees or keyword lists, and leave the model the freedom to choose how to attain the desired quality levels.

Enabling the direct control of the three quality dimensions, allows flexibility with respect to the specific requirements of the task at hand, and opens a range of generation possibilities: paraphrases of various flavors (e.g. syntactically vs. lexically diverse), quasi-paraphrases (with lower semantic similarity), and even non-paraphrases which may be useful for downstream tasks (e.g. hard negative examples of sentences that are linguistically similar but have different meanings \cite{guo-etal-2018-effective, Reimers2020MakingMS}).

Our results show that the \textsc{QCPG} model indeed enables controlling paraphrase quality along the three quality dimensions.

Furthermore, even though the training data is of mixed quality, and exhibits scarcity in the high quality area (see Figure \ref{fig:data_density_plot}), our model is able to learn high quality paraphrasing behavior, i.e. it increases the linguistic diversity of the generated paraphrases without decreasing the semantic similarity compared to the uncontrolled baseline.

\section{Method}
\label{sec:method}
\begin{figure*}
\includegraphics[width=\textwidth]{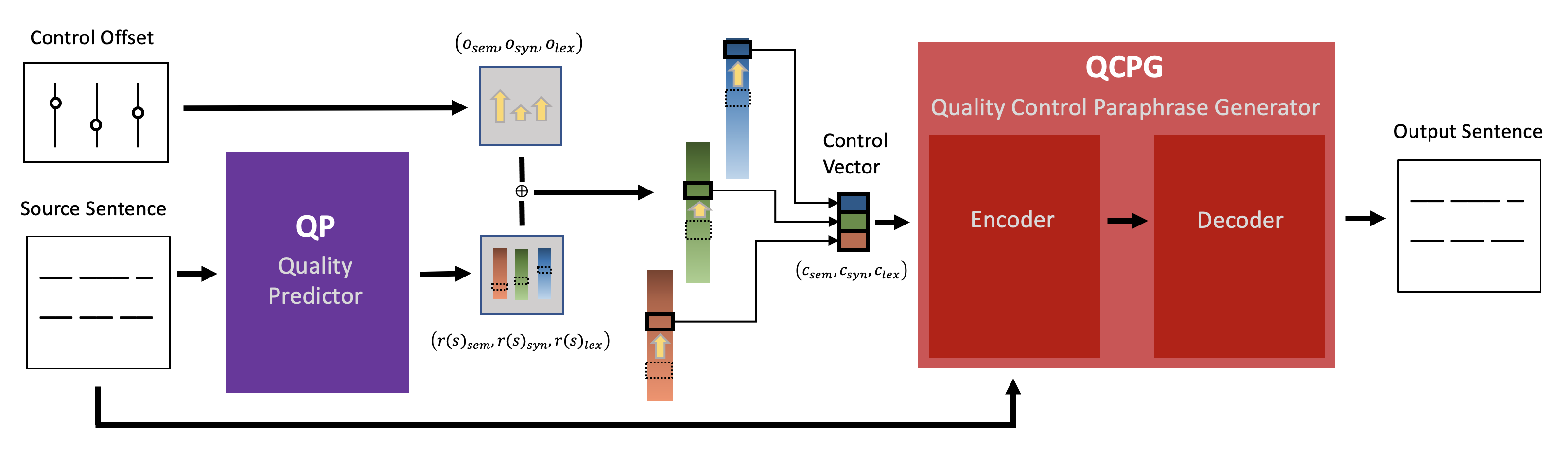}
\caption{Solution Architecture. 
The input to the paraphrase generation model, QCPG, is composed of two elements: a sentence $s$, and a three-dimensional quality vector $c = (c_{sem}, c_{syn}, c_{lex})$, which controls the quality of the generated paraphrase. Selecting appropriate values of $c$ is crucial for obtaining high-quality paraphrases.
The quality predictor model, QP, helps select suitable input quality vectors, by predicting the typical quality, $r(s)$, of the  paraphrases of $s$. The control vector $c$ is the sum of $r(s)$, and an offset vector $o$, which indicates the extent to which the requested quality deviates  from the typical value. Dev-set results can help the user in selecting suitable  values of $o$, as shown in Figure \ref{fig-heatmap}} 

%The input is composed of two elements, a sentence $s$, and a  control offset $\bf{o}=(o_{sem},o_{syn},o_{lex})$. 
%The QP model predicts the quality reference point $\bf{r}(s)$.
%The input vector to the \textsc{QCPG} model, which controls the quality of the generated paraphrase of $s$, is the sum of the reference point $\bf{r}(s)$ and the offset $\bf{o}$.
%The input control vector to the \textsc{QCPG} model, which generates the paraphrase of $s$, is the sum of the reference point and the offset.
%The quantized point now can get embeddings for every quantized axis with additional quality. The text and the quality added vector embeddings are fed into the encoder-decoder which generate the resulting text. 
%}
\label{fig-arch}
\end{figure*}

In this section we provide a general description of our approach.
We first explain how the different quality dimensions are measured. We then describe the controlled paraphrase generation model, \textsc{QCPG}, and finally we suggest a method that given the task requirements, detects the input control values which 
maximize the quality of the generated paraphrases.
Figure \ref{fig-arch} summarizes our proposed solution for generating controlled paraphrases, which is detailed in the rest of the section.

\subsection{Quantifying Paraphrase Quality}
\label{quality metrics}
The most common dimensions for measuring paraphrase quality are the semantic, syntactic and lexical dimensions. Several previous works used also a fluency evaluation metric \cite{siddique2020unsupervised}. However, since our focus is on the supervised setting, we rely on the gold paraphrases as fluency guidance for the model \cite{McCarthy2009TheCO}. Thus, given a sentence $s$ and a paraphrase $s'$, we define the paraphrase quality as a three dimensional vector $\mathbf{q}(s,s') = (q_{sem}(s,s'),q_{syn}(s,s'),q_{lex}(s,s'))$,
where $q_{sem}$ is a measure of semantic similarity, and $q_{syn}$ and $q_{lex}$ are measures of syntactic and lexical variation, respectively. 
For the syntactic score, inspired by \citet{iyyer-etal-2018-adversarial-aug} we choose $q_{syn}(s,s')$ to be the normalized tree edit distance \cite{zhang1989simple} between the third level constituency parse-trees of $s$ and $s'$, after removing the tokens - to increase the decoupling from the lexical distance metric. 
% Removing the tokens helps decouple the syntactic diversity measure from the lexical one.
We define the lexical score $q_{lex}(s,s')$ to be the normalized character-level minimal edit distance between the bag of words. 
This measure is independent of word order, and hence increases the decoupling from syntactic measures.
Additionally, calculating the token distances on the character level enables to capture tokens that share the same stem/lemma.
 Character-level distance is also more robust to typos that may be found in noisy data.
As for the semantic score, several strong metrics have been recently proposed for measuring semantic similarity between sentences.
In order to select $q_{sem}(s,s')$, we studied the agreement between the candidate metrics and human judgments, using only development data, and found Bleurt \cite{sellam2020bleurt} to have the highest correlation with human judgments (see Appendix \ref{semantic metric selection}). Thus, we define $q_{sem}(s,s')$ to be the Bleurt score, normalized using the sigmoid function to ensure a uniform range of values, $[0,1]$, for all three quality dimensions.
For ease of presentation all metrics are presented on a $0-100$ scale.

\subsection{The \textsc{QCPG} Model} 
The main component of our solution is a quality controlled paraphrase generation model (\textsc{QCPG}), which is an encoder-decoder model trained on the task of controlled paraphrase generation.
Given an input sentence $s$ and a control vector $\mathbf{c} = (c_{sem},c_{syn},c_{lex})$, the goal of \textsc{QCPG} is to generate an output paraphrase $QCPG(s,\mathbf{c})$ that conforms to $\mathbf{c}$. 
We train \textsc{QCPG} using the training set pairs $(s,t)$, by setting $\mathbf{c}$ to be $\mathbf{q}(s,t)$, and maximizing  $P(t|s,\mathbf{c}=\mathbf{q}(s,t))$ over the training set via the autoregressive cross entropy loss.

\subsection{Control Values Selection}
\label{Control Values Selection}
A major challenge in the research of controlled paraphrase generation, is selecting appropriate input control values
 that can be achieved by the model \cite{goyal2020neural}.
Clearly, given a sentence, not all paraphrase qualities are achievable. 
Some sentences are more amenable to paraphrasing than others. For example, named entities and numbers are much harder to be replaced while keeping sentence meaning, 
and hence, the potential lexical diversity of paraphrases involving such terms is relatively limited.
Forcing \textsc{QCPG} to conform to quality control values that are too high with respect to the input sentence, may lead to suboptimal quality of the resultant paraphrases.
Thus, for a more effective use of \textsc{QCPG}, the control values should be determined with respect to the input sentence. 

Below we describe the second part of our solution, namely a method that given a sentence, predicts the input control values, $\mathbf{c}(s)$, that optimize the expected quality of the paraphrases generated by \textsc{QCPG}.
For simplicity we assume that the quality distribution $p(\mathbf{q}|s)$ of all paraphrases of sentence $s$, is approximately normally distributed around a sentence dependent mean $\mathbf{q_0}(s)$, and that the variance is approximately sentence-independent.
We further assume that given an input sentence $s$, the difficulty to generate a paraphrase of a given quality, $\mathbf{q}$, is dominated by 
$p(\mathbf{q}|s)$ rather than by the quality vector $\mathbf{q}$ itself. 

Following our assumptions, the level of difficulty can be expressed by the offset, $\mathbf{o}=(o_{sem},o_{syn},o_{lex})$ of $\mathbf{q}$ from $\mathbf{q_0}(s)$. 
Thus, the input control, $\mathbf{c}(s)$, for \textsc{QCPG}, is the sum of $\mathbf{q_0}(s)$ and an offset $\mathbf{o}$.

Our aim is to analyze the model results for varying levels of difficulty, namely under different offsets, $\mathbf{o}$, from $\mathbf{q_0}(s)$. 

\textbf{The Quality Predictor (\textsc{QP}):} Since $\mathbf{q_0}(s)$ is unknown, we introduce \textsc{QP}, a regressor whose output, termed the reference of $s$, $\mathbf{r}(s)=(r_{sem}(s),r_{syn}(s),r_{lex}(s))$, approximates $\mathbf{q_0}(s)$. 
During training, \textsc{QP} aims to predict $\mathbf{q}(s,t)$ given $s$, where $(s,t)$ are the input-output pairs of the training data.

To summarize, we define \textit{sentence-aware quality control} by decomposing the \textsc{QCPG} input control, $\mathbf{c}$, into a sum of a sentence dependent reference point, $\mathbf{r}(s)$, and a sentence independent offset, $\mathbf{o}$. 
\section{Data and Implementation Details}
\label{sec:details}

\subsection{Datasets}
To test the ability of our model to learn high quality behavior from mixed quality data we use weakly annotated datasets. These datasets are  large but noisy, and contain only a relatively small amount of high quality paraphrases.

\textbf{MSCOCO}: This dataset consists of 123K images, where each image contains at most five human-labeled captions \cite{Lin2014MicrosoftCC}.
 Similar to previous works 
 %on paraphrase generation, 
 we consider different captions of the same image as paraphrases. 

% The MSCOCO dataset contains image captions of 123K images with five captions per image, assuming that captions associated with the same picture are paraphrases. Each source sentence is associated with four ground truth paraphrases. 
% \elron{Even though MSCOCO used for evaluation of previous paraphrasing system it contains many very weak paraphrases}

\textbf{WikiAnswers (WikiAns for short)}: The WikiAnswers corpus  contains clusters of questions tagged by wiki-answers.com users as similar. There are $30,370,994$ clusters with $25$ question in each on average. 
In total, the corpus
contains over $70$ million question pairs \cite{Fader14}. 

\textbf{ParaBank2.0}: A dataset containing clusters of sentential paraphrases, produced from a bilingual corpus using negative constraints, inference sampling, and clustering \cite{hu-etal-2019-large}. The dataset is composed of avarage of $5$ paraphrases in every cluster and close to $100$ million pairs in total.

To get comparable results across all datasets, we randomly sub-sampled ParaBank2.0 and WikiAns to the same size as MSCOCO, and split them to train, dev and test sets, of sizes $900K$, $14K$ and $14K$ respectively. We carefully made sure that there are no pairs from the same cluster in different splits of the data. The full data splits will be published with our code.

\subsection{Implementation Details}
%All models mentioned in the paper are trained with batch size of $32$ on $2$ A100 for $6$ epochs. Full details as well as train and dev results can be found in Appendix \ref{models}.
All models are trained with batch size of $32$ on $2$ NVIDIA A100 GPUs for $6$ epochs. Full details as well as train and dev results can be found in Appendix \ref{models}.

\textbf{\textsc{QCPG:}}
We use the pre-trained T5-base \cite{T5} as the encoder-decoder model.
% on a mixture of unsupervised and supervised tasks, each of which is converted into a sequence-to-sequence format. 
The control input vector to \textsc{QCPG} is quantized at every dimension into $20$ equally spaced values ranging from $0$ to $100$. Each value is assigned to a special saved-token.
The three tokens corresponding to the quantized values of the control vector $\bf{c}$, are concatenated to the head of the input sentence, and together used as input to the model.  $\bf{r(s)}$ and $\bf{o}$ are also quantized in a similar way. 

\textbf{QP:}
An Electra base model~\cite{clark2020electra} finetuned with MSE loss to predict the typical quality values (see Section \ref{Control Values Selection}).

\textbf{Baseline Model (BL):} A T5-base model finetuned on the training data.

 For all the models, we adopt the experimental setup used in \citep{devlin-etal-2019-bert}, i.e. we train the model with several learning rates and choose the one that achieves the highest dev set performance (see appendix \ref{models}).
%For each dataset, the uncontrolled model , which is used as a baseline, is T5-base, finetuned on the training data.

\section{Results}

\subsection{Controlling the Quality Dimensions}
\label{Controlling the Quality Dimensions}

The aim of the following analysis is to study the level of control achieved by \textsc{QCPG}. To this end, we measure the model response to changes in the input offsets. We compute
the expected difference in paraphrase quality, as a result of applying an input offset $\mathbf{o}$ compared to zero offset as a reference.
More formally, we define the 3-dimensional responsiveness vector of \textsc{QCPG} at an offset $\mathbf{o}$, $\mathbf{R}(\mathbf{o})$ as $\mathbf{Q}(\mathbf{o})-
\mathbf{Q}((0,0,0))$,
 where $\mathbf{Q}(\mathbf{o})$ is the expected quality of the paraphrases generated by \textsc{QCPG} at an offset $\mathbf{o}$. We estimate $\mathbf{Q(\mathbf{o})}$ by averaging $\mathbf{q}(QCPG(s,\mathbf{r}(s)+\mathbf{o}))$ over the input sentences $s$ of the dev set, and denote this estimate by $\mathbf{\tilde{Q}(\mathbf{o})}=(\mathbf{\tilde{Q}_{sem}(\mathbf{o})},\mathbf{\tilde{Q}_{syn}(\mathbf{o})},\mathbf{\tilde{Q}_{lex}(\mathbf{o})})$, and the corresponding estimate of $\mathbf{R}(\mathbf{o})$ by $\mathbf{\tilde{R}(\mathbf{o})}$.

Specifically, in the following analysis we are interested in studying the model response to each of the dimensions separately, i.e. how changing the input offset along a given quality dimension $dim$ -- the \textit{controlled} dimension -- while keeping the two other dimensions constant, affects the responsiveness in each of the three dimensions.
A good control mechanism would imply that increasing the input offset in one dimension will result in a monotonically increasing responsiveness in that dimension, with relatively small responsiveness in the other two dimensions.

\begin{figure*}
%\vspace{-1cm}
%\includegraphics[width=0.95\linewidth]{control.pdf}
\includegraphics[width=0.95\linewidth]{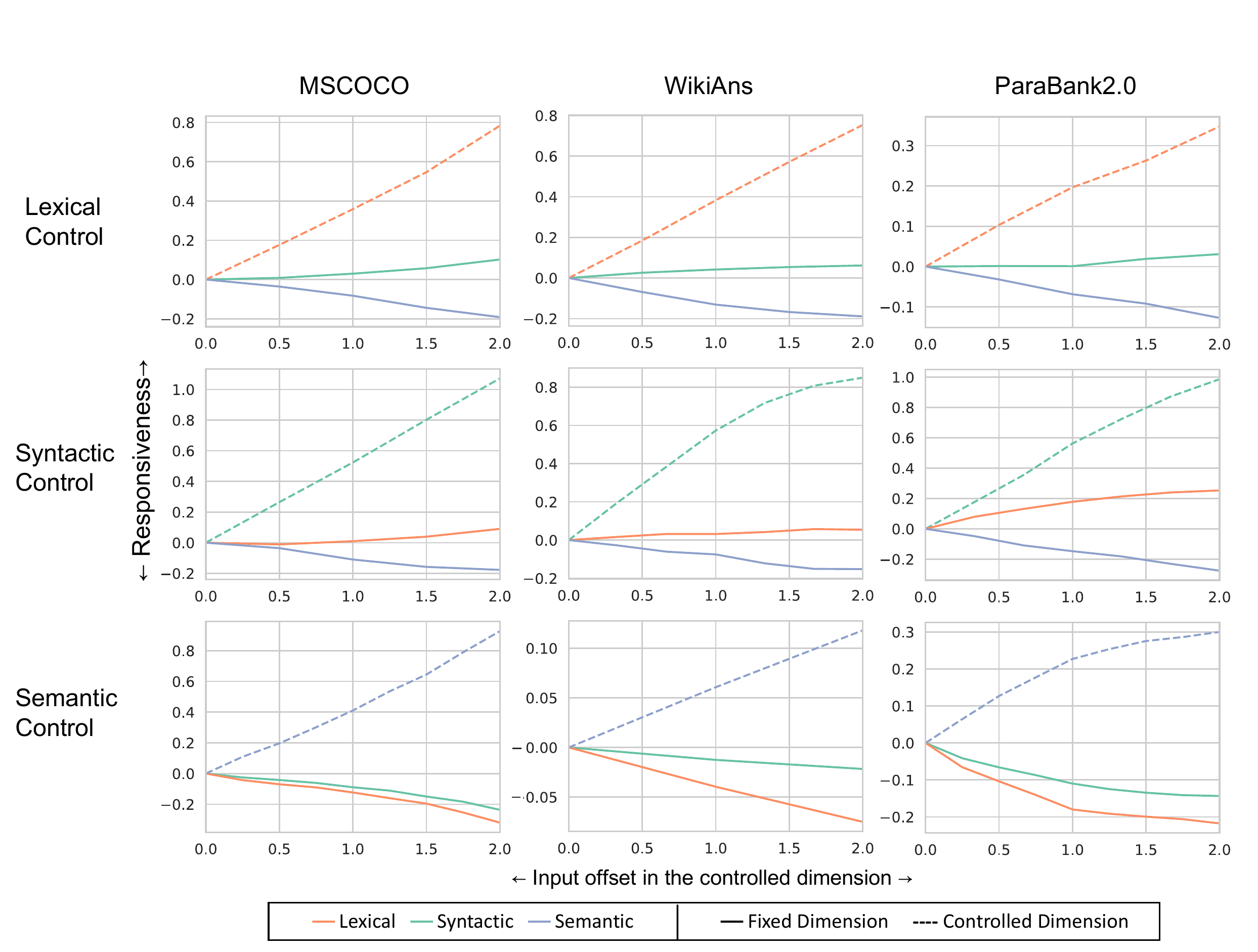}
\caption{
Controllability of \textsc{QCPG}. The  responsiveness of \textsc{QCPG} to changes in the input quality vector. In each graph only one dimension of the input is changed (the control dimension), where the other two dimensions are fixed at zero offset. The control dimensions in the top middle and bottom rows are the lexical syntactic and semantic dimensions respectively. Each color represents a different quality dimension of the generated paraphrases. The responsiveness in the control dimension is plotted in a dashed line.
Responsiveness and offsets are shown in standard deviation units.
%plotted for the two datasets. 
%Responsiveness is averaged over the dev set.
% Dashed line: the responsiveness in the controlled dimension.}
%\caption{Control over separate dimensions of the paraphrasing. At every plot we controlled one attribute and kept the others stagnant and report the output offset from the origin at each dimension in the plot lines.
}
\label{fig-control}
%\vspace{-1cm}
\end{figure*}

Figure \ref{fig-control} shows, for each of the three datasets,  the responsiveness in the three quality dimensions, when changing the input offset along each of the three dimensions, while fixing the input offsets in the other two dimensions at $0$. 
Examining the actual values of quality in the paraphrases of the dev sets, reveals that the standard deviation is different in each dimension. Hence, for clarity of presentation, we present the input offset values and the responsiveness in units of standard deviation as measured in the respective dimension and dev set.  
  
For the range of offsets displayed in Figure \ref{fig-control}, the responsiveness in the controlled dimension increases monotonically with the input offsets across all datasets and dimensions.
As expected, the responsiveness in the uncontrolled dimensions does not zeros due to the inherent coupling between the dimensions. 
For example, many changes that increase syntactic diversity, also increase lexical diversity 
(e.g. a move from passive to active voice). 
% In reality, the quality dimensions are coupled, to some extent, for various reasons. For example, many changes that increase syntactic diversity, also increase lexical diversity 
% by definition 
% (e.g. a move from passive to active voice). 
Still, our control mechanism is able to increase the responsiveness in the controlled dimension with relative low responsiveness in the uncontrolled dimensions. 
%As expected, the dimensions are coupled, to some extent, due to the inherent dependency between them. 
%Still, using our control mechanism, the responsiveness in the uncontrolled dimensions is much smaller than in the controlled dimension. 
Specifically, focusing on the relation between semantic similarity and expression diversity, the figure shows that there is a minor decrease in semantic similarity in response to an increase in lexical and syntactic diversity. In the next section, we will show that this does not prevent our model from generating paraphrases that are not only more lexically and syntactically diverse, but also more semantically similar to the source sentences, compared to the paraphrases generated by the uncontrolled baseline.

\begin{figure*}
%\vspace{-1cm}
\begin{center}
\includegraphics[width=1.0\linewidth]{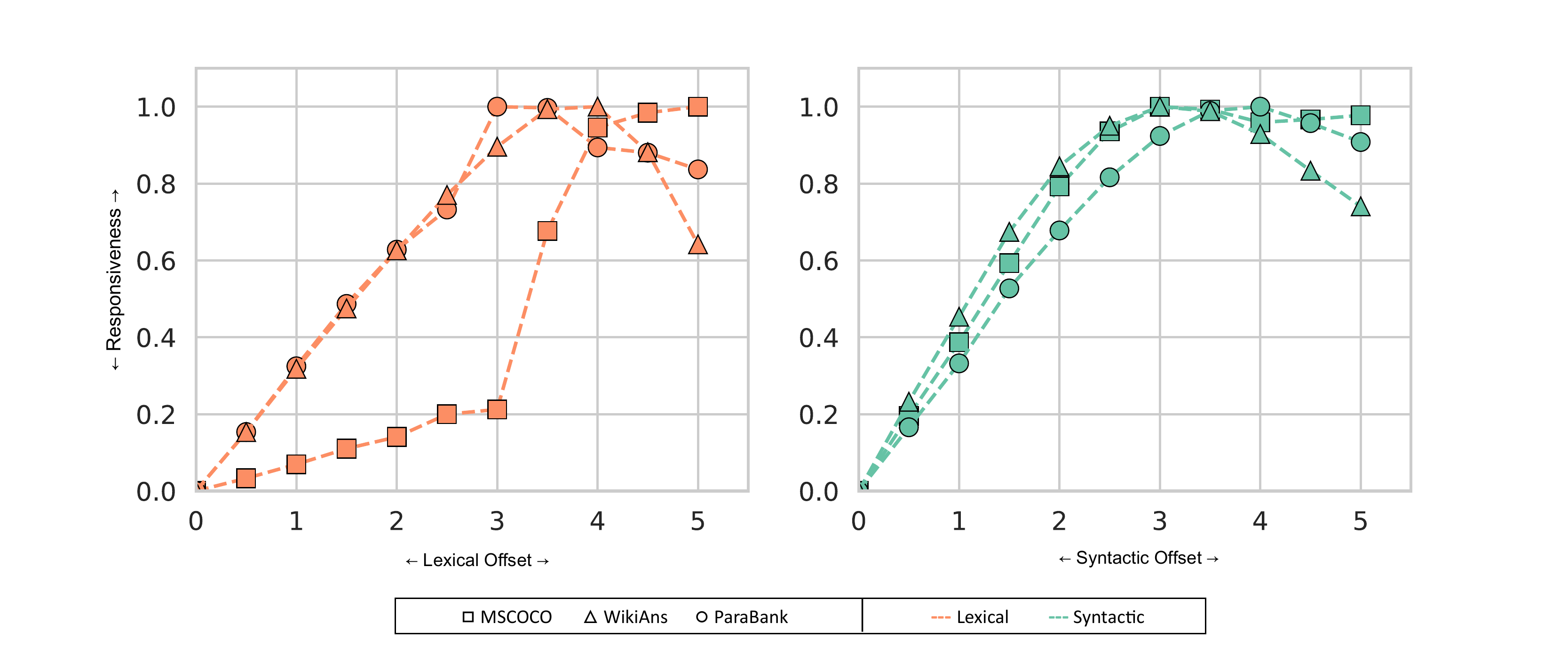}

\caption{Monotonicity break in the responsiveness of \textsc{QCPG}. 
The  responsiveness of \textsc{QCPG} in the controlled quality dimension vs. the  input offset in that dimension in the three datasets. Left: Lexial control. Right: Syntactic control. 
Each curve is normalized by its maximal value, to create a uniform y-axis range across all datasets.
Offsets are shown in standard deviation units.}
% when looking at the extended range we can see the breaking of monotonoticity \elron{should be integrated: "Since our main interest here is to demonstrate the \textit{pattern} of behavior of the responsiveness curve across datasets, rather than its actual \textit{values}, for the sake of presentation, we normalize each curve by its maximal value, to create a uniform y-axis range across all datasets, without affecting the shapes of the curves."}}
%\caption{Control over separate dimensions of the paraphrasing. At every plot we controlled one attribute and kept the others stagnant and report the output offset from the origin at each dimension in the plot lines.}
\label{fig-control-extended}
\end{center}
\end{figure*}

Figure \ref{fig-control} focused on small to moderate input offsets, i.e. offsets up to $2$ stds from the reference point. 
However, as we speculated before, with increasing offsets, i.e. the more the requested control value deviates from the typical value, it becomes increasingly difficult to generate a paraphrase that conforms to the requested control value.
Figure \ref{fig-control-extended} depicts the responsiveness in the syntactic and lexical dimensions for a larger range of offset values.
For the semantic dimension, the  typical values are too high to allow large positive offsets, which for most sentences result in exceeding the upper limit of the semantic score.
Indeed, as can be seen in Figure~\ref{fig-control-extended}, when moving to high offset values, the responsiveness in the syntactic and lexical dimensions starts to decrease. 
This behavior is in line with our aforementioned hypothesis, and reflects the detrimental effect of feeding \textsc{QCPG} with input control values that are too far from the typical paraphrase qualities of the input sentence. 
The non-monotonic behavior of the responsiveness implies that the input offsets should be selected carefully in order to optimize the quality of the resultant paraphrases. In Section~\ref{sec:control-quality} we suggest a method for identifying these optimal offsets.

\subsection{Selecting Optimal Input Control Values}
\label{sec:control-quality}
%We analyze the effect of the control values on the resulting paraphrases. The effect is measured using two key quantities: the linguistic diversity (the average of the syntactic and lexical distance -- see Methods) and the semantic similarity
 
In this section, we suggest a method that given task requirements, selects the input offsets that are expected to yield the desired quality of paraphrases.
% In the previous section, we focused on how changing the input offset vector in a single dimension from the all-zero vector affects the \textit{responsiveness} of the controlled dimension and its \textit{decoupling} from the other dimensions. 
% Here, we analyze the effect of the input control values on the \textit{actual quality} of the resulting paraphrases, not only along a single dimension, but when concomitantly controlling all 3-dimensions.
The idea is to compute the estimated expected quality, $\mathbf{\tilde{Q}(o)}$,
% the dev set in order to estimate 
for each input offset $\mathbf{o}$, using the dev set as described in Section~\ref{Controlling the Quality Dimensions},
% the expected quality of the paraphrases generated by \textsc{QCPG} at this offset,
and then search the 3D grid of input offsets to find the point
for which $\mathbf{\tilde{Q}(\mathbf{o})}$ is best suited for the user's requirements.
%that corresponds to the optimal $\mathbf{\tilde{Q}(o)}$ under the user constraints.
We envision this analysis as a preliminary step in which the user chooses the input control parameters that best achieve his desired paraphrasing operation point, and then uses the chosen values at inference -- which is why we use the dev set. 
% Notice that this analysis does not require further labeled data.

 %Similarly, the expected quality vector of the uncontrolled baseline is estimated by averaging $\mathbf{q}(BL(s))$, over the input sentences of the dev set, where $BL(s)$ is the paraphrase of the input sentence $s$, generated by the uncontrolled baseline model.
%Unlike in the previous section, where the offsets were studied along each dimension separately, here we study the full 3-dimensional grid of offset points in the relevant range, i.e every $\mathbf{o}$ where $o_{sem}$, $o_{syn}$ and $o_{lex}$ in ${0,5,10...40}$.

\begin{figure*}

\begin{center}
    \includegraphics[width=1\linewidth]{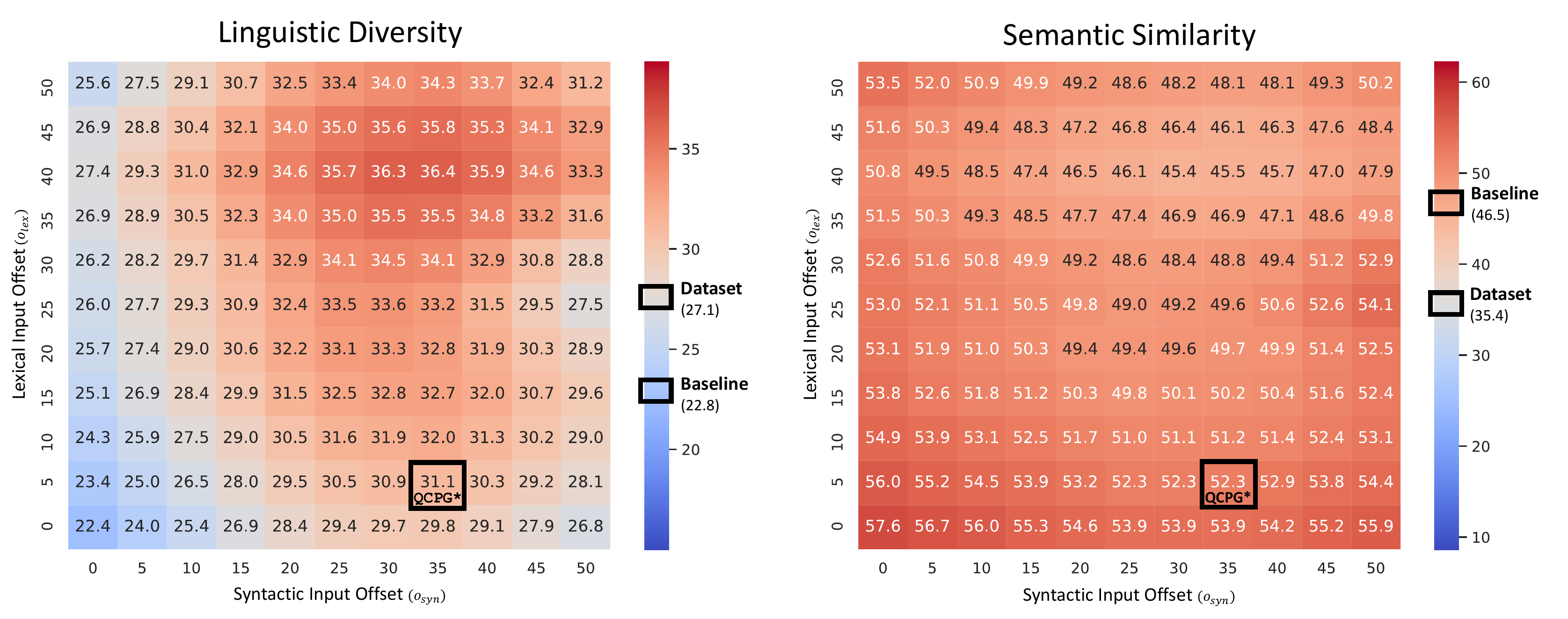}

\caption{Estimated Quality at different offset values for WikiAns. 
Average of  linguistic diversity (left) and semantic similarity (right) of the paraphrases generated for the dev-set sentences, as a function of $o_{syn}$ and $o_{lex}$, for fixed $o_{sem}=50$.
The average quality of the gold-label paraphrases, and the average values achieved by the uncontrolled baseline, are marked on the color bars.
%Average is over the dev set. 
%The number in each grid point is the difference from the average value over the dev set. 
Red/blue shades correspond to above/below the dev-set mean. 
%Top three rows: COCO dataset with $o_{sem}=0,5,10$. Bottom row: QQP with $o_{sem}=5$. 
}
\label{fig-heatmap}
\end{center}
\end{figure*}
We study the behavior of $\mathbf{\tilde{Q}(o)}$ as a function of the 3D grid of offset points in the relevant range, i.e every $\mathbf{o}$ where $o_{sem}$, $o_{syn}$ and $o_{lex}$ in ${0,5,10...50}$. 
Figure \ref{fig-heatmap} depicts $\mathbf{\tilde{Q}(\mathbf{o})}$ for WikiAns, on a slice of the full offset grid.
The results for the full grid on all datasets are shown in Figure \ref{heatmaps}.
The right-hand-side map depicts the estimated linguistic diversity (the average of $\mathbf{\tilde{Q}_{syn}}(\mathbf{o})$ and $\mathbf{\tilde{Q}_{lex}}(\mathbf{o})$) and the left-hand-side depicts the semantic similarity, $\mathbf{\tilde{Q}_{sem}}(\mathbf{o})$). The maps are presented for $o_{sem}=50$, and for different values of $o_{syn}$ and $o_{lex}$. 
As expected, the two measures are anti-correlated, where areas with increased semantic similarity are characterized by decreased linguistic diversity. 
The \textsc{QCPG} results are compared to two reference points, which are invariant to $\mathbf{o}$ and are marked on the colorbars with black squares: 'Dataset' is the semantic-similarity/linguistic-diversity average value over the corresponding dev set paraphrases, and 'Baseline' is the average semantic-similarity/linguistic-diversity of the uncontrolled baseline over the corresponding dev set.
Notice that the average diversity level achieved by the uncontrolled baseline is lower than that of the dev set mean, reflecting the difficulty of this model to generate diverse paraphrases. 
\textsc{QCPG} on the other hand, with suitable input offset values, is able to generate paraphrases which are on average higher than the baseline both in their linguistic diversity and in their semantic similarity, and in fact even higher in many cases than the values of the ground truth paraphrases in the dev-set. 
%Different input offset values lead to different output qualities, many of which outperform the baseline and in many cases, also the dataset ground truth performance. 
%There are plenty points that are better than the baseline. 
%As can be seen in Figure \ref{fig-heatmap} there are plenty such points that are better than the baseline. In the following section we exemplify the  capabilities of our model by choosing one of many good points and evaluating it.

In general, the estimates of the expected quality achieved by \textsc{QCPG} at different input offsets, enable a user to generate paraphrases at different operation points, by manipulating the input offset control $\mathbf{o}$ to meet her desired quality values.
Consider for example a typical use case, of aiming to maximize linguistic diversity under a constraint on semantic similarity.
An example of such a case is an operation point, denoted by \QCPGPoint, which aims to exemplify the advantage of \textsc{QCPG} over the baseline, by
maximizing linguistic diversity under the constraint that the semantic similarity is at least $5$ points higher than the baseline.
%Consider for example an operation point, denoted by \QCPGPoint, which corresponds to the maximal linguistic diversity that can be achieved under the constraint that the semantic similarity level is at least $5$ points higher than the baseline. The exact offset values to obtain this operation point depend on the dataset. 
The input offset values to obtain this operation point depend on the dataset, and can be found using heatmaps such as in Figure \ref{fig-heatmap}. 
For WikiAns the input offset for the \QCPGPoint \space operation point values are $(50,35,5)$ (entry marked by the black square).

\subsection{Quality Evaluation on the Test Set}
% We now turn to evaluate our model using the test sets. 
% We perform this evaluation on the \QCPGPoint operation point, though it can be extended to other operation points. 
% For each test set, we identify the corresponding offsets using the methodology described in Section \ref{sec:control-quality}.
% %We use the source sentences of the test set to evaluate the quality of the paraphrases generated by \QCPGPoint. 
% %Specifically, we use \QCPGPoint as our operation point, and identify the corresponding offsets using the methodology described in Section \ref{sec:control-quality}. 
% Since the quality estimates are based only on the dev set, we can safely use them for  selecting suitable offsets when evaluating the model on the test set. 
% We first perform an evaluation using automatic quality measures, and then report results based on human judgments.  The results are compared to the uncontrolled baseline.

In the previous section we saw, using estimates based on the dev sets, that there are many operation points which generate paraphrases with higher quality than those achieved by the uncontrolled baseline. 
We now turn to evaluate one such operation point, namely \QCPGPoint, using the source sentences of the \textit{test} sets which were not used in the selection of the input offset values.

%\subsubsection{Automatic Evaluation}
\paragraph{Automatic Evaluation}
%We use four quality metrics for evaluating different aspects of the paraphrases generated by our model: sem, syn, lex and i-BLUE. sem, syn and lex are the quality metrics we used for the control of $QCPG$, and are described in Section \ref{quality metrics}. Additionally, we use i-BLUE \cite{sun2012joint} by following the metrics in recent work \cite{li2019decomposable,liu2019unsupervised}, which aims to measure the linguistic diversity in the generated paraphrases by penalizing copying words from input sentences.
We use four quality measures to evaluate different aspects of generated paraphrases. 
%sem, syn and lex are the quality measures used in the control of $QCPG$ (Section \ref{quality metrics}). 
The three quality measures used in the control of \textsc{QCPG} (Section \ref{quality metrics}) and Self-BLEU \cite{zhu2018texygen} as adapted in  \citet{li2019decomposable,liu2019unsupervised}, which aims to measure the linguistic diversity in the generated paraphrases by penalizing copying from input sentences.
%Additionally, we use i-BLEU \cite{sun2012joint} by following the metrics in recent work \cite{li2019decomposable,liu2019unsupervised}, which aims to measure the linguistic diversity in the generated paraphrases by penalizing copying words from input sentences.
As can be seen in Table \ref{tab:AutomaticEvaluation}, \QCPGPoint \space outperforms the baseline in all metrics across all datasets, as predicted using the dev-set heatmaps. 
A clear advantage is obtained even for Self-BLEU, which was not part of the metrics used as input controls. 
Importantly, the quality of the paraphrases generated by our model is comparable to, or at times better than the quality of the paraphrases in the ground truth of the datasets. Examples of paraphrases generated by \QCPGPoint \space  compared to the ground truth paraphrases appear in Table \ref{tab:generation}. This is an important step towards the goal of obtaining paraphrases in the sparse area of high quality (recall the top right corner of Figure \ref{fig:data_density_plot}). 

% \elron{Another perspective of evaluation is the  effect of the quality guidance on the model's ability to predict the ground truth paraphrases.}
Additionally, we examined \textsc{QCPG} from another perspective: the  effect of the quality guidance on the model's ability to predict the ground truth paraphrases. Tables \ref{tab:baseline_train} and \ref{tab:qcpg_train} show the BLEU scores \cite{papineni-etal-2002-bleu} obtained by \textsc{QCPG} and the uncontrolled baseline respectively. The results verify that the input quality vectors induced by the target sentences are effectively utilized by  \textsc{QCPG} to achieve better prediction performance.  
% Finally, BLEU score \cite{papineni-etal-2002-bleu} results show that  the model with the quality dimensions of the target paraphrases, help better at predicting the ground-truth paraphrase targets compared to the uncontrolled baseline (see Tables \ref{tab:qcpg_train} and \ref{tab:baseline_train})  

%  our results show that our method is better at predicting the ground-truth paraphrase targets as measuered by BLEU score  \ref{tab:qcpg_train} \ref{tab:baseline_train}.

% We further examine the usefulness of the control mechanism from another perspective - the ability of the controlled model to predict the target paraphrases. For this purpose we used the ground-truth target sentences to measure the average BLUE score of \QCPGPoint \space (see Table \ref{tab:qcpg_train})  compared to the baseline model (see Table \ref{tab:baseline_train}). The higher BLUE scores obtained by \QCPGPoint \space verify the ability of the model to utilize the quality information provided in the input for better prediction of the target sentences. 
%  In order to evaluate our model performance, we used three operation points, $sem+$, $syn+$ and $lex+$, representing three different user preferences.
% Given a dimension $x$, the point $x+$ maximizes the average of the estimated quality over the other two dimensions,  under the condition that the estimated quality in $x$ is at least $5$ points above the baseline.
% WRITE ABOUT THE RESULTS
% REMEMBER TO STRESS THAT THE OPERATION POINTS WERE CHOSEN USING DEV AND HERE WE LOOK AT TEST AND SEE THAT INDEED WE GET WHAT WE EXPECTED

\begin{table*}[htbp]
\centering
\resizebox{\linewidth}{!}{
\begin{tabular}{l l c c c c| c c c c |c c c c}
\toprule
 & & \multicolumn{4}{c}{MSCOCO} & \multicolumn{4}{c}{WikiAns} & \multicolumn{4}{c}{ParaBank2} \\
 \midrule
  & & $q_{sem}\uparrow$ & $q_{syn}\uparrow$ & $q_{lex}\uparrow$& \small{Self-BLEU}$\downarrow$& $q_{sem}\uparrow$& $q_{syn}\uparrow$& $q_{lex}\uparrow$& \small{Self-BLEU}$\downarrow$& $q_{sem}\uparrow$& $q_{syn}\uparrow$& $q_{lex}\uparrow$& \small{Self-BLEU}$\downarrow$\\
 \hline 
  Gold & & 29.9 & \underline{34.5} & 28.0 & \underline{8.7} & 34.6 & 30.7 & 24.4 &
\underline{16.4} & 75.0 & 18.5 & \underline{20.9} & \underline{23.9}  \\
  \hline
 BL & & 50.0 & 27.8 & 23.0 & 18.8 & 46.6 & 24.7 & 20.9 & 23.4 & 77.8 & 16.8 & 18.6	 & 29.4  \\
\QCPGPoint{} & & \underline{\textbf{56.6}} & \textbf{29.6} & \underline{\textbf{42.4}} & \textbf{18.0} & \underline{\textbf{48.5}} & \underline{\textbf{41.5}} & \underline{\textbf{24.8}} & \textbf{21.4} & \underline{\textbf{81.4}} & \underline{\textbf{18.9}} & \textbf{19.6} & \textbf{27.1}  
 \\
 \bottomrule
\end{tabular}}
\caption{Automatic evaluation of the \textsc{QCPG} model on the test set. The semantic similarity ($q_{sem}$), syntactic diversity ($q_{syn}$)  and lexical diversity ($q_{lex}$), are measured using Bleurt, Tree edit distance, and character-level edit distance respectively, as described in Section \ref{sec:method}. 
Self-BLUE is an external  measure of linguistic diversity (see text for details).
BL: uncontrolled baseline. Gold: the test set ground truth paraphrases. 
\QCPGPoint{} is the \textsc{QCPG} model in the operation point defined in Section \ref{sec:control-quality}. Best performance amongst the compared models is highlighted in \textbf{bold}. Best results amongst the models and the gold labels are underlined.}
\label{tab:AutomaticEvaluation}

\end{table*} 

\begin{table}[!]
\centering
\resizebox{1.0\columnwidth}{!}{
\begin{tabular}{l c c c c}
\toprule
& \multicolumn{3}{c}{Votes} & Agreement\\
\midrule
 & \QCPGPoint{} & BL & (Tie) & Cohen’s Kappa \\ \hline
MSCOCO & \textbf{.56} & .36 & (.08) &  .38\\ 
WikiAns & \textbf{.48} & .36 & (.16) &  .47\\ 
ParaBank2 & \textbf{.30} & .26 & (.44) & .57\\ 
\bottomrule 
\end{tabular}
}
\caption{Human evaluation of semantic similarity. The numbers represent the proportion of annotators that voted for each method. \QCPGPoint{}: the \textsc{QCPG} model in the operation point defined in Section \ref{sec:control-quality}. BL: Uncontrolled Baseline. % Tie: 'both' or no decision.
}
\label{tab:SemanticHuman}
\end{table} 
\vspace{-0.1cm}
\paragraph{Human Evaluation} 
While linguistic diversity can be automatically measured by reliable metrics such as Self-BLEU, measuring semantic similarity is more challenging. 
We therefore rely on automatic metrics for evaluating the lexical and syntactic diversity, but use human annotation for validating the semantic evaluation.   
%In addition to the automatic evaluation, 
To this end, we selected a sample of $50$ source sentences from each test set, and generated one paraphrase using the uncontrolled baseline and one using \QCPGPoint. 
%We then obtained judgments from Appen\footnote{\url{appen.com}} crowd workers. 
%For each of the datasets, 
The annotators were shown the source sentence, along with the two generated paraphrases (randomly ordered), and were asked which of the two better 
preserves the semantic meaning of the source sentence
% serves as a paraphrase to the original 
(ties are also allowed). 
% For half of the original sentences the paraphrases were generated with the Baseline model, and the \textsc{QCPG} models, and the for the other half with the Baseline model and by XX\michal{Liat, could you please describe here how you generated the 'TQ'}. 
In total, $150$ triplets were evaluated by $5$ judges. %, $50$ triples per dataset. 
%Table~\ref{tab:SemanticHuman} summarizes the majority vote, as well as the inter-annotator agreement, which was measured using Cohen's Kappa \cite{mchugh2012interrater}.
Table~\ref{tab:SemanticHuman} demonstrates an advantage for \QCPGPoint \space in all datasets, with a large margin in MSCOCO and WikiAns. This advantage is statistically significant ($p-value<0.05$) as obtained by applying the Wilcoxon signed-rank test to the difference between the number of annotators that voted for \QCPGPoint \space and those voted for the baseline, across all datasets. 
Thus, the human evaluation is in line with the results of the automatic semantic similarity measure. We also verified,
that the results of this sample, in terms of linguistic diversity, are very similar to those shown in Table \ref{tab:AutomaticEvaluation}.

For examples of paraphrases generated by \QCPGPoint  \space see Table  \ref{tab:generation} in the Appendix.

\section{Related Work}
\label{Related work}
 Many recent works on paraphrase generation have been focused on attempting to achieve  high-quality 
 paraphrases. These works can be divided into supervised and unsupervised approaches. 

\textbf{Supervised Approaches}
% Two main directions were explored for improving quality in supervised paraphrase generation: data-oriented and model-oriented. In the data-oriented methods, the main idea is to find automatic ways to create large-scale data with linguistic diversity, in order to improve the quality of paraphrases generated by the subsequent models.  \cite{dong2021parasci} created a paraphrase dataset in the scientific domain based on characteristics and common patterns of scientific papers. In \cite{lan2017continuously}, 
% large-scale paraphrases were collected from
% Twitter by linking tweets through shared URLs.
% Another line of works leveraged Neural Machine Translation (NMT),
% and increased diversity by adding diversity-oriented elements to the NMT decoding procedure \cite{hu2019large}. 
% The main drawback of the data-oriented  approaches is that they are based on domain-specific properties/resources such as shared-urls or parallel data, which may not be applicable to low-resource domains.
% Among the model-oriented approaches, 
To achieve diversity, some works focused on diverse decoding using heuristics such as Hamming
distance or distinct n-grams to preserve diverse options during beam search  \cite{vijayakumar2018diverse}. Other works generate multiple outputs by perturbing latent representations
 \cite{gupta2018deep,park2019paraphrase}.
or by using distinct
generators \cite{qian2019exploring}. These
methods achieve some diversity, but do not control generation in an interpretable manner.

The works that are most similar to ours strive to gain diversity using  controlled-paraphrase generation, 
by exposing control mechanisms that are manipulated to produce either lexically \cite{zeng2019user,thompson-post-2020-paraphrase} or syntactically \cite{chen2019controllable,goyal2020neural} diverse paraphrases.
One approach is to use an exemplar sentence for guiding the syntax of the generated paraphrase   \cite{chen2019controllable,bao2019generating, hosking-lapata-2021-factorising}. An alternative is to directly employ constituency tree as the syntax guidance \cite{iyyer-etal-2018-adversarial-aug,li-choi-2020-transformers}. \citet{goyal2020neural} promote syntactic diversity by conditioning over possible syntactic rearrangements of the input. 
\citet{zeng2019user} use keywords as lexical guidance for the generation process.
Here we introduce a simple model for jointly controlling the lexical, syntactic and semantic aspects of the generated paraphrases.  

\textbf{Unsupervised Approaches}
 \citet{niu2020unsupervised} rely on neural models to generate high quality paraphrases, using a decoding method that enforces diversity by preventing repetitive copying of the input tokens. 
\citet{liu-etal-2020-unsupervised} optimize a quality oriented objective by casting  
paraphrase generation as an optimization problem,
and searching the sentence space to find the optimal point. 
\citet{garg2021unsupervised} and \citet{siddique2020unsupervised} use reinforcement learning with quality-oriented reward combining textual entailment, semantic similarity, expression diversity and fluency. In this work, we employ similar metrics for guiding the generation of paraphrases within the \textit{supervised} framework.

\section{Discussion}
In this paper, we propose a novel controlled paraphrase generation model, that leverages measures of paraphrase quality for encouraging the generation of paraphrases with desired quality. We demonstrate the high level of control achieved by the model, and suggest a method for coping with the challenging problem of finding suitable control values.

Aside from offering a simple and effective way for controlling models' output quality, the quality control paradigm enables a holistic view of
the data, the training process and the final model analysis.
%(\RomanNumeralCaps{1}) the data (\RomanNumeralCaps{2}) the training process and (\RomanNumeralCaps{3}) the final model analysis.
Namely: 
(\RomanNumeralCaps{1}) Examination of the training data through the lens of data quality enables to characterize the data at hand, its strengths and limitations. 
(\RomanNumeralCaps{2}) A quality-aware training process can be viewed as multi-task learning, where each quality level is a separate task with its own accurate supervision, as opposed to the standard quality-agnostic approach, where low quality data is in fact used as a poor supervision for a model which aims at generating higher quality output. (\RomanNumeralCaps{3}) Analyzing the model behavior under different quality controls, allows finer understanding of the different model behaviors and the trade-offs between their output qualities.
Better understanding the expected output quality of neural NLG models, for different input quality controls, can increase the trust in their output.
 
Finally, our model analysis consistently shows that although the models generally follow the quality requirements, there is still room for improvement. 
A possible direction for future research is exploring methods, such as reinforcement learning, for further improving the ability of the model to satisfy the quality requirements.

% A future research direction can focus on the enforcement of the quality controlled satisfaction with methods such as reinforcement learning.    

% Our analysis shows that there is a gap between the quality consitions and the actual quality achieved by the model output.
% show that when increasing quality controls, the output quality monotonically growing. we also show that there is more room for improvements in making the model conform to the actual value of quality required future works can address this issue by enforcing that the quality demands accurately satisfied by the model. 

% Our analysis show that our model is able to only approximately follow the quality conditions
% In this work, we assume that the paraphrase generation task must take into account not only the question of whether a sentence is a paraphrase of another, but also what type of paraphrase. We use paraphrase quality information to characterize existing datasets, define control measures, and develop a new generation model.

\bibliographystyle{acl_natbib}
\bibliography{anthology, acl2021}

\appendix
% \section{Appendix}

\section{Selecting the semantic similarity measure}
\label{semantic metric selection}
Recently, several strong metrics have been proposed for measuring semantic similarity between sentences \cite{ reimers-gurevych-2019-sentence, zhang2019bertscore, sellam2020bleurt}. In order to select the semantic similarity metric for \textsc{QCPG}, we performed a small experiment over the three dev sets, with the aim of measuring the agreement of the candidate metrics with human judgments.
To this end, we leveraged two properties that characterize weakly labeled datasets, the underlying clusters of sentences, and the high variability of semantic similarity. Given a dataset, we randomly selected $100$ clusters, and picked three sentences at random from each cluster. 
For each triplet of sentences $t=(t_1,t_2,t_3)$ we asked $5$ human annotators to choose which of the two sentences, $t_2$ or $t_3$, better preserves the semantic meaning of $t_1$. 
In order to find the candidate similarity measure with the highest agreement with human judgments, 
we first computed, for each triplet, the difference between the number of annotators voted for $t_2$ and those voted for $t_3$. 
We then computed for each candidate measure, the difference between the similarity of $t_2$ to $t_1$ and and of $t_3$ to $t_1$. 
We then measured Kendall's Tau correlation \cite{daniel1990applied} between the difference vector of the human judgments and that of the judgments of each of the candidate measures. 
Table \ref{tab:corr_human} shows the resultant correlations. The highest correlations are obtained for SBERT \cite{reimers-gurevych-2019-sentence}, but since it was trained on WikiAns and MSCOCO, we could not use it in our study.  
% The results appear in the Appendix \ref{}. 
We selected Bleurt due to its highest correlation with human judgments over the three datasets (among the methods that were not exposed to the considered datasets). We normalize Bleurt score using the sigmoid function to ensure a uniform range of values, $[0,1]$, for the three quality dimensions. 

\section{Correlation of semantic similarity measures with linguistic diversity}
\label{sem sim corr with ling div}
We study the coupling between the different semantic similarity measures and the linguistic diversity. We assume that the level of coupling of a good similarity measure will resemble that of humans, and will be less sensitive to lexical and syntactic properties of the paraphrase. Table \ref{tab:corr_diversity} presents the Kendall tau correlation between the different similarity measures and the linguistic diversity. Results for human judgments are also shown for a reference.  The correlation calculation is performed between the vectors of differences as described in section \ref{semantic metric selection}). The results show that Bleurt demonstrates the lowest coupling with linguistic diversity among the automatic measures (aside from SBERT which, as mentioned before, was trained with MSCOCO and WikiAns). The comparison to human judgments shows that Bleurt is more influenced by linguistic features, indicating that automatic measures need to be further improved to reach the decoupling level achieved by humans. 
\begin{table}
\centering
\resizebox{0.95\columnwidth}{!}{%

        \begin{tabular}{p{2.8cm} l r r}
    \toprule
              & MSCOCO &  WikiAns &  ParaBank2 \\
     \midrule
     SBERT     &   .52  &    .43 &      .41 \\
     \midrule
     BERTSCORE &   .38  &    .3 &      .31 \\
     BLEURT    &   \textbf{.45}  &        \textbf{.4}    &      \textbf{.36} \\

    \bottomrule
    \end{tabular}
}
\caption{Correlation of different semantic similarity models  with human evaluations.}
\label{tab:corr_human}
\end{table}

\begin{table}
\centering
\resizebox{0.95\columnwidth}{!}{%

        \begin{tabular}{p{2.8cm} r r r}
    \toprule
              & MSCOCO &  WikiAns &  ParaBank2 \\

     \midrule
     Human     & -0.17   &   -0.19 &      -0.25 \\
     \midrule
     SBERT     &   -0.17  &   -0.37 &      -0.29 \\
     \midrule
     BERTSCORE &   -0.39  &    -0.48 &      -0.51 \\
     BLEURT    &   \textbf{-0.25}  &        \textbf{-0.36}    &      \textbf{-0.39} \\

    \bottomrule
    \end{tabular}
}
\caption{Correlation of different semantic similarity models with linguistic diversity.}
\label{tab:corr_diversity}
\end{table}

\section{Models Details and Training Results} The learning rates for the \textsc{QCPG} and the Baseline models were selected in the following way. For a given dataset, we finetuned the models with $4$ learning rates ($1e$-$3$, $1e$-$4$, $5e$-$3$, $5e$-$4$) (The training results of the baseline presented in Table \ref{tab:baseline_train} and the results of QCPG presented in Table \ref{tab:qcpg_train}.). For the baseline we selected the one which yielded the best BLEU score \cite{papineni-etal-2002-bleu} on the corresponding dev set The best learning rate for every dataset was chosen based on the Dev set BLEU score. For the \textsc{QCPG}  we chose the model that best conforms to the control input as measured by the MSE between the input control vector and the output quality vector (see Table \ref{tab:qcpg_mse}).
The \textsc{QP} model is an Electra-Base model finetuned with $4$ different learning rates ($1.5e$-$4$, $1e$-$4$, $3e$-$5$, $5e$-$5$). We choose the learning rate the yields the minimal MSE on the dev set (For full results see Table \ref{tab:qp_train})

\subsection{Full Heatmaps}
\begin{figure*}

\begin{center}
    \includegraphics[width=0.9\linewidth]{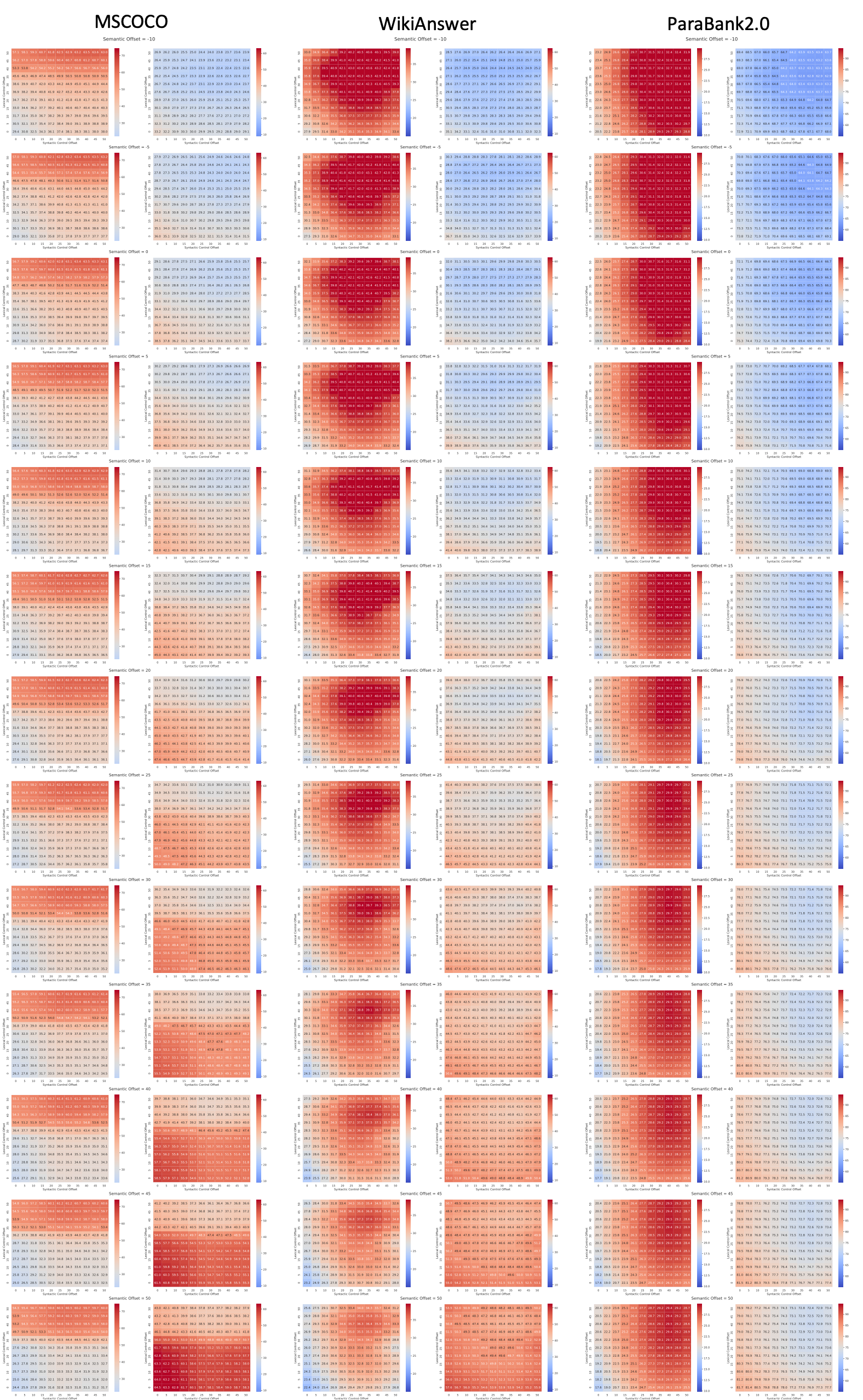}

\caption{Heatmaps of linugstic diversity (left column) and semantic similarty (right column) as a function of input control offsets for the datasets.}
\label{heatmaps}
\end{center}
\end{figure*}

The full heatmaps can be found in Figure \ref{heatmaps}.
\label{models}
\begin{table}
\centering
\resizebox{0.95\columnwidth}{!}{
\begin{tabular}{llrrr}
\toprule
     Dataset  &     LR &  Dev BLEU $\uparrow$ &  Dev Loss $\downarrow$ &  Train Loss  \\
\midrule
 \multirow{4}{*}{MSCOCO} & 1e-3 &    10.19 &     2.10 &      1.52 \\
        & 1e-4 &    \textbf{10.94} &     \textbf{1.89} &      1.65 \\
        & 5e-3 &     0.00 &     2.23 &      2.76 \\
        & 5e-4 &    10.53 &     2.07 &      1.51 \\
        \midrule
\multirow{4}{*}{ParaBank2} & 1e-3 &    27.28 &     1.38 &      0.65 \\
         & 1e-4 &    \textbf{30.22} &     \textbf{1.15} &      0.69 \\
         & 5e-3 &     0.00 &     3.45 &      3.88 \\
         & 5e-4 &    28.40 &     1.37 &      0.61 \\
         \midrule
\multirow{4}{*}{WikiAns} & 1e-3 &    13.09 &     2.24 &      1.46 \\
         & 1e-4 &    \textbf{15.22} &     \textbf{1.95} &      1.59 \\
         & 5e-3 &     0.00 &     3.62 &      4.03 \\
         & 5e-4 &    13.51 &     2.17 &      1.43 \\
\bottomrule

\end{tabular}
}

\caption{Training and dev set loss of the finetuned T5 baseline.}
\label{tab:baseline_train}
\end{table}
\begin{table}
\centering
\resizebox{0.95\columnwidth}{!}{%
\begin{tabular}{llrrr}
\toprule
   Dataset     &     LR &  Dev BLEU &  Dev Loss &  Train Loss \\
\midrule
 \multirow{4}{*}{MSCOCO} & 1e-3 &      11.14 &       2.01 &        1.47 \\
        & 1e-4 &      11.24 &       1.80 &        1.61 \\
        & 5e-3 &       0.00 &       2.29 &        2.89 \\
        & 5e-4 &      10.86 &       1.98 &        1.46 \\
\midrule
\multirow{4}{*}{ParaBank2} & 1e-3 &      32.03 &       1.28 &        0.60 \\
        & 1e-4 &      34.28 &       1.05 &        0.65 \\
        & 5e-3 &       0.00 &       3.37 &        3.86 \\
        & 5e-4 &      32.77 &       1.25 &        0.56 \\
\midrule
\multirow{4}{*}{WikiAns} & 1e-3 &      17.29 &       2.08 &        1.40 \\
        & 1e-4 &      19.48 &       1.81 &        1.52 \\
        & 5e-3 &       0.00 &       3.57 &        4.01 \\
        & 5e-4 &      18.21 &       1.99 &        1.36 \\
\bottomrule

\end{tabular}
}
\caption{Training and dev set loss of the QCPG.}
\label{tab:qcpg_train}
\end{table}
\begin{table}
\centering
\resizebox{0.95\columnwidth}{!}{%
\begin{tabular}{lrrrr}
\toprule
Dataset &  Diversity &  Lexical &  Syntactic &  Semantic \\

\midrule
MSCOCO  &             25.4 &           23.0 &           27.8 &          50.0 \\
ParaBank2 &             17.7 &           18.6 &           16.8 &          77.8 \\
WikiAns &             22.8 &           20.9 &           24.7 &          46.6 \\
\bottomrule
\end{tabular}}
\caption{Automatic evaluation of the chosen finetuned T5 baseline.}
\end{table}

\begin{table}
\centering
\resizebox{0.95\columnwidth}{!}{%

        \begin{tabular}{p{2.8cm} l r r}
    \toprule
    Dataset &       LR &  Dev MSE $\downarrow$&  Train MSE \\
    \midrule
     \multirow{4}{*}{MSCOCO} & 1.5e-4 &    0.0242 &      0.0240 \\
             &   1e-4 &    0.0242 &      0.0240 \\
              &   3e-5 &    0.0206 &      0.0161 \\
              &   5e-5 &    \textbf{0.0205} &      0.0164 \\
              \midrule
    \multirow{4}{*}{ParaBank2} & 1.5e-4 &    0.0260 &      0.0239 \\
             &   1e-4 &    0.0248 &      0.0239 \\
             &   3e-5 &    \textbf{0.0169} &      0.0124 \\
             &   5e-5 &    0.0170 &      0.0126 \\
             \midrule
    \multirow{4}{*}{WikiAns} & 1.5e-4 &    0.0402 &      0.0374 \\
             &   1e-4 &    0.0404 &      0.0374 \\
             &   3e-5 &    \textbf{0.0317} &      0.0200 \\
             &   5e-5 &    0.0445 &      0.0372 \\
    \bottomrule
    \end{tabular}
}
\caption{Training results of the QP models.}
\label{tab:qp_train}
\end{table}

\begin{table}
\centering
\resizebox{1\columnwidth}{!}{%
\begin{tabular}{p{3.2cm} p{2.8cm} l}
\toprule
Dataset & LR &       MSE $\downarrow$ \\
\midrule
\multirow{4}{*}{MSCOCO} & 1e-3 & 0.0124 \\
        & 1e-4 & 0.0119 \\
        & 5e-3 & 0.2943 \\
        & 5e-4 & \textbf{0.0118} \\
\multirow{3}{*}{ParaBank2} & 1e-3 & 0.0140 \\
        & 1e-4 & 0.0129 \\
        & 5e-4 & \textbf{0.0125} \\
\multirow{4}{*}{WikiAns} & 1e-3 & 0.0166 \\
        & 1e-4 & \textbf{0.0153} \\
        & 5e-3 & 0.3091 \\
        & 5e-4 & 0.0155 \\
\bottomrule
\end{tabular}
}
\caption{MSE between the required control and the evaluations of the outputs of the QCPG models.}
\label{tab:qcpg_mse}
\end{table}

\begin{table*}
\centering
\resizebox{0.95\linewidth}{!}{%
\def\arraystretch{1.25}\tabcolsep=10pt
\begin{tabular}{p{5cm}  p{5cm}  p{5cm}}

  MSCOCO \\
   \toprule
  
                                                   Source &  Ground-truth &                             \QCPGPoint \\
    \midrule
    A table filled with assorted prepared foods in a buffet fashion.&	Fresh fruits, vegetables, and other foods are spread out on the table. &	A table with food on it in a buffet line. \\
    Ornately decorated assortment of vases displayed on shelf. &	A display of pottery in a glass case &	A decorated shelf with vases on display \\
    Group of people seated at a long table eating pizza &	A group of people are sitting around a wooden table.&	A group of people sitting at a long table with pizza. \\
    A building with a clock and weather vane is outlined against the blue sky. &	a building with a clock inside of it &	A clock and weather vane on a blue sky. \\
    A knitted teddy bear hanging off an afghan &	A blue crocheted teddy bear hanging off of a crocheted blanket &	A knitted teddy bear hanging from a quilt \\
    Two men pose next to a huge vase with an owl painted on it. &	a big vase sits in the middle of a couple of people &	Two men standing next to a large vase with an owl on it. \\
    
    \bottomrule
\\
  WikiAns \\
  \toprule
  
                                                   Source &  Ground-truth &                             \QCPGPoint \\
   \midrule
What did the cheyennes indians do for a living? &	Cheyenne indians live in the derest? &	What kind of jobs did the Cheyenne Indians have? \\
% Can you get a mini bike on GTA san andreas? &	How do you get a poloni pit bike on gta san andreas on ps2? &	How do you get a mini bike in gta san andreas for ps2? \\
What temperature scale do you use in australia? &	Temperature scale used for scientific work? &	What is the temperature scale for Australia? \\
Are there any other names for tay sachs disease? & 	Who is warren tay and bernard sachs? &	Other names for tay sachs disease?  \\
What should you give to your elder sister on her birthday?	& What should you get your little sister for her 9th birthday? &	Your older sister's birthday what to give? \\
How changes in the respiration rate affect blood pH? &	How does Increase in respiration of water affect pH? &	Explain how the respiration rate affects the pH? \\
What is the value of a dollar bill signed by joseph w barr? &	What is the value of a dollar bill 1963 signed by joseph barr? &	Joseph W Barr dollar bill value? \\
What are the three meninges that cover the brain and spinal cord? &	The three memebranous coverings that protect the brain and spinal cord? &	What three meninges cover the brain and spinal cord? \\
                 
\bottomrule
\\
  ParaBank2 \\
  \toprule
  
                                                   Source &  Ground-truth &                             \QCPGPoint \\
                                                   \midrule
We're having trouble with Roger.&	I've got issues on Roger.&	We have a problem with Roger. \\
Everything on schedule.	& All on schedule. &	All in the plan. \\
The internet no longer maked the distance matter: the world may indeed be our classroom. &	Because of the Internet, distance doesn't matter anymore: the world may indeed be an our classroom. & 	The Internet doesn't matter: the world could be our school. \\
Article 2 deals with the scope of application of a directive extending cooperation between Member States to include taxes of whatever type.&	Article 2 concerns an area which is covered by a Directive which broadens cooperation among Member States so as that it covers taxes of any kind.&	Article 2 concerns the scope of the directive extending the cooperation between the Member States to include taxation of any kind. \\
You're free to move forward. &	You're free to move on.	& You can go on.\\
\bottomrule
\end{tabular}}
\caption{Paraphrases generated by \QCPGPoint  compared to ground-truth paraphrases.}
\label{tab:generation}
\end{table*}

\end{document}